\documentclass{llncs}
\usepackage{makeidx}  

\usepackage{cite}
\usepackage{amssymb}  
\usepackage{amsmath}
\DeclareMathOperator{\dist}{dist}
\usepackage[pdftex]{graphicx}
\graphicspath{{./images/}{./images/}}
\DeclareGraphicsExtensions{.pdf,.jpeg,.png,.jpg}
\usepackage{subfig}
\usepackage{url}
\usepackage{hyperref} 

\begin{document}
\mainmatter              
\title{Building Networks for Image Segmentation using Particle Competition and Cooperation}
\titlerunning{Building Networks for Image Segmentation using Particle Competition and Cooperation}  
%
\author{Fabricio Breve}
\authorrunning{Fabricio Breve} 
%
\tocauthor{Fabricio Breve}
\institute{S\~{a}o Paulo State University (UNESP), Rio Claro SP 13506-900, Brazil,\\
\email{fabricio@rc.unesp.br},\\ WWW home page:
\texttt{http://www.rc.unesp.br/igce/demac/fbreve/}
}

\maketitle              

\begin{abstract}
Particle competition and cooperation (PCC) is a graph-based semi-supervised learning approach. When PCC is applied to interactive image segmentation tasks, pixels are converted into network nodes, and each node is connected to its k-nearest neighbors, according to the distance between a set of features extracted from the image. Building a proper network to feed PCC is crucial to achieve good segmentation results. However, some features may be more important than others to identify the segments, depending on the characteristics of the image to be segmented. In this paper, an index to evaluate candidate networks is proposed. Thus, building the network becomes a problem of optimizing some feature weights based on the proposed index. Computer simulations are performed on some real-world images from the Microsoft GrabCut database, and the segmentation results related in this paper show the effectiveness of the proposed method.
\keywords{particle competition and cooperation, image segmentation, complex networks}
\end{abstract}
\section{Introduction}
Image Segmentation is the process of dividing an image into multiple parts, separating foreground from background, identifying objects, or other relevant information \cite{Shapiro2001}. This is one of the hardest tasks in image processing \cite{Gonzalez2008} and completely automatic segmentation is still a big challenge, with existing methods being domain dependent. Therefore, interactive image segmentation, partially supervised by an specialist, became an interesting approach in the last decades \cite{Boykov2001,Grady2006,Protiere2007,Blake2004,Ducournau2014,Ding2010,Rother2004,Paiva2010,Li2010,Artan2010,Artan2011,Xu2008}.

Many interactive image segmentation approaches are based on semi-supervised learning (SSL), category of machine learning which is usually applied to problems where unlabeled data is abundant, but the process of labeling them is expensive and/or time-consuming, usually requiring intense work of human specialists \cite{Zhu2005,Chapelle2006}. SSL techniques employ both labeled and unlabeled data in their training process, overcoming the limitations of supervised and unsupervised learning, in which only labeled or unlabeled data is used for training, respectively. Regarding the interactive segmentation task, SSL techniques spread labels provided by the user for some pixels to the unlabeled pixels, based on their similarity.

Particle competition and cooperation (PCC) \cite{Breve2012TKDE} is a graph-based SSL approach, which employs particles walking on a network represented by an undirected and unweighted graph. Nodes represent the data elements and the particles represent the problem classes. Particles from the same class cooperate with each other and compete against particles representing different classes for the possession of the network nodes.

Many graph-based SSL techniques are similar and share the same regularization framework \cite{Zhu2005}. They usually spread the labels globally, while PCC employs a local propagation approach, through the walking particles. Therefore, its computational cost is close to linear ($O(N)$) in the iterative step, while many other state-of-the-art methods have cubic computational complexity ($O(N^3)$).

PCC was already applied to some important machine learning tasks, such as overlapped community detection \cite{Breve2013SoftComputing,Breve2011ISNN}, learning with label noise \cite{Breve2010IJCNN,Breve2015Neurocomputing,Breve2012SBRN}, learning with concept drift \cite{Breve2012IJCNN,Breve2013BRICS-ConceptDrift}, active learning \cite{Breve2013IJCNN,Breve2013BRICS-ActiveLearning,Breve2014ENIAC}, and interactive image segmentation \cite{Breve2015IJCNN,Breve2015ICCSA}.

In the interactive segmentation task, PCC is applied to a network built from the image to be segmented. Each pixel is represented by a network node. Edges are created between nodes corresponding to similar pixels. Then, particles representing the labeled pixels walk through the network trying to dominate most of the unlabeled pixels, spreading their label and trying to avoid invasion from enemy particles representing other classes in the nodes they already possess. In the end of the iterative process, the particles territory frontiers are expected to coincide with the frontiers among different image segments \cite{Breve2015IJCNN}.

In the network formation stage, the edges between nodes are created based on the similarity between the corresponding pixels, according to the Euclidean distance between their features, which are extracted from the image. A large amount of features may be extracted from each pixel. These include RGB (red, green, and blue) components, intensity, hue, and saturation. Other features take pixel location and neighborhood into account. Given an image, each feature may have more or less discriminative capacity regarding the classes of interest. Therefore, it is important to weight each feature according to its discriminative capacity, so the PCC algorithm segmentation capacity is also increased.

Unfortunately, defining these weights is a difficult task. The methods proposed so far work well in some images, but fail in others. In \cite{Breve2015WVC}, four automatic feature weight adjustment methods were proposed based on feature values (mean, standard deviation, histogram) for each class in the labeled pixels. They were applied to three images from the Microsoft GrabCut database \cite{Rother2004}. Three of the methods were able to increase PCC segmentation accuracy in at least one image, but none of them increased accuracy in all of the three images.

In this paper, a new method to automatically define feature weights is proposed. It is based on an index, which is extracted from candidate networks built with all the features and their candidate weights. This approach has some advantages over the methods that consider only individual features. For instance, individual features may not be good discriminators, but combined they may have a higher discriminative capacity. A candidate network is built considering the combination of all features and their respective weights. Therefore, the proposed index, extracted from the candidate networks, may be used to evaluate if a given set of weights leads to a proper network to be used by PCC.

In this sense, finding a good set of weights is just a matter of optimizing the weights based on the index extracted from the candidate network built with them. In this paper, a genetic algorithm \cite{Goldberg1989,Mitchell1998} is used to optimize the weights, with the proposed index used as the fitness function to be maximized.

Computer simulations are performed using some real-world images extracted from the Microsoft GrabCut database. The PCC method is applied to both a network built with feature weights optimized by the proposed method and a network built with non-weighted features, used as baseline. The segmentation accuracy is calculated on both resulting images, comparing them to the \emph{ground truth} images labeled by human specialists. The results show the efficacy of the proposed method.

The remaining of this paper is organized as follows. Section \ref{sec:particlemodel} presents the particle competition and cooperation model. In Section \ref{sec:pesos}, the proposed method is explained. Section \ref{sec:experiments} presents the experiments used to validate the method. In section \ref{sec:results}, the computer simulation results are presented and discussed. Finally, some conclusions are drawn on Section \ref{sec:conclusions}.

\section{Image Segmentation using Particle Competition and Cooperation}
\label{sec:particlemodel}

In this section, the semi-supervised particle competition and cooperation approach for interactive image segmentation is presented. The reader can find more complete expositions in \cite{Breve2015IJCNN} and \cite{Breve2012TKDE}.

Overall, PCC may be applied to image segmentation tasks by converting each image pixel into a network node, represented by an undirected and unweighted graph. Edges among nodes are created between similar pixels, according to the Euclidean distance between the pixel features. Then, a particle is created for each labeled node, i.e., nodes representing labeled pixels. Particles representing the same class belong to the same team, they cooperate with their teammates to dominate unlabeled nodes, at the same time that they compete against particles from other teams. As the system runs, particles walk through the network following a random-greedy rule.

Each node has a set of domination levels, each level belonging to a team. When a particle visits a node, it raises its team domination level on that node, at the same time that it lowers the other teams domination levels. Each particle has a strength level, which changes according to its team domination level on the node its visiting. Each team of particles also has a table to store the distances between all the nodes it has visited and the closest labeled node of its class. These distance tables are dynamically updated as the particles walk. At the end of the iterative process, each pixel will be labeled by the team that has the highest domination level on its corresponding node.

A large amount of features may be extracted from each pixel $x_i$. In this paper, $23$ features are considered: (1) the pixel row location; (2) the pixel column location; (3) the red (R) component of the pixel; (4) the green (G) component of the pixel; (5) the blue (B) component of the pixel; (6) the hue (H) component of the pixel; (7) the saturation (S) component of the pixel; (8) the value (V) component of the pixel; (9) the ExR component; (10) the ExG component; (11) the ExB component; (12) the average of R on the pixel and its neighbors (MR); (13) the average of G on the pixel and its neighbors (MG); (14) the average of B on the pixel and its neighbors (MB); (15) the standard deviation of the R on the pixel and its neighbors (SDR); (16) the standard deviation of G on the pixel and its neighbors (SDG); (17) the standard deviation of B on the pixel and its neighbors (SDB); (18) the average of H on the pixel and its neighbors (MH); (19) the average of S on the pixel and its  neighbors (MS); (20) the average of V on the pixel and its neighbors (MV); (21) the standard deviation of H on the pixel and its neighbors (SDH); (22) the standard deviation of S on the pixel and its neighbors (SDS); (23) the standard deviation of V on the pixel and its neighbors (SDV).

For all measures considering the pixel neighbors, an $8$-connected neighborhood is used, except on the borders where no wraparound is applied. All components are normalized to have mean $0$ and standard deviation $1$. They may also be scaled by a vector of weights $\lambda$ in order to emphasize/deemphasize each feature during the network generation. ExR, ExG, and ExB components are obtained from the RGB components using the method described in \cite{Lichman2013}. The HSV components are obtained from the RGB components using the method described in \cite{Smith1978}.

The network is represented by the undirected and unweighted graph $\mathbf{G} = (\mathbf{V},\mathbf{E})$, where $\mathbf{V} = \{v_1,v_2,\dots,v_N\}$ is the set of nodes, and $\mathbf{E}$ is the set of edges $(v_i, v_j)$. Each node $v_i$ corresponds to the pixel $x_i$. Two nodes $v_i$ and $v_j$ are connected if $v_j$ is among the $k$-nearest neighbors of $v_i$, or vice-versa, considering the Euclidean distance between the features of $x_i$ and $x_j$. Otherwise, $v_i$ and $v_j$ are disconnected.


For each node $v_i \in \{v_1,v_2,\dots,v_L \}$, corresponding to a labeled pixel $x_i \in \mathfrak{X}_{L}$, a particle $\rho_i$ is generated and its initial position is defined as $v_i$. Each particle $\rho_j$ has a variable $\rho_j^\omega(t) \in [0, 1]$  to store its strength, which defines how much it impacts the node it is visiting. The initial strength is always set to maximum, $\rho_j^{\omega}(0)=1$.

Each team of particles has a distance table, shared by all the particles belonging to the team. It is defined as $\mathbf{d_c(t)} = d_c^1(t),\dots,d_c^N(t)\}$. Each element $d_c^i(t) \in [0 \quad N-1]$ stores the distance between each node $v_i$ and the closest labeled node of the class $c$. Particles initially know only that the distance to any labeled node of its class is zero ($d_c^i=0$ if $y(x_i)=c$). All other distances are adjusted to the maximum possible value ($d_c^i=n-1$ if $y(x_i) \neq c$) and they are updated dynamically as the particles walk.

Each node $v_i$ has a dominance vector $\mathbf{v_i^\omega(t)} = \{v_i^{\omega_1}(t), v_i^{\omega_2}(t), \dots, v_i^{\omega_C}(t) \}$, where each element $v_i^{\omega_c}(t) \in [0, 1]$ corresponds to the domination level of the team $c$ over the node $v_i$. The sum of all domination levels in a node is always constant:
\begin{equation}
    \sum_{c=1}^{C} v_i^{\omega_c} = 1.
\end{equation}

Nodes corresponding to labeled pixels have constant domination levels, and they are always adjusted to maximum for the corresponding team and zero for the others. On the other hand, nodes that correspond to unlabeled pixels have variable dominance levels, initially equal for all teams, but varying as they are visited by particles. Therefore, for each node $v_i$, the dominance vector $\mathbf{v_i^\omega}$ is defined by:
\begin{equation}\label{eq:NodesInit}
    v_i^{\omega_c}(0) = \left\{
    \begin{array}{ccl}
        1 & & \mbox{if $x_i$ is labeled and $y(x_i) = c$} \\
        0 & & \mbox{if $x_i$ is labeled and $y(x_i) \neq c$} \\
        \frac{1}{C} & & \mbox{if $x_i$ is unlabeled}
    \end{array}\right..
\end{equation}

When a particle $\rho_j$ visits an unlabeled node $v_i$, domination levels are adjusted as follows:
\begin{equation}\label{eq:UpdateNodesPot}
    v_i^{\omega_c}(t+1) = \left\{
    \begin{array}{l}
        \max\{0,v_i^{\omega_c}(t) - \frac{0,1 \rho_j^{\omega}(t)}{C-1}\} \\
        \quad \mbox{if $c \neq \rho_j^c$} \\
        v_i^{\omega_c}(t) + \sum_{r \neq c}{v_i^{\omega_r}(t)-v_i^{\omega_r}(t+1)} \\
        \quad \mbox{if $c = \rho_j^c$} \\
    \end{array}\right.,
\end{equation}
where $\rho_j^c$ represents the class label of particle $\rho_j$. Each particle $\rho_j$ will change the node its visiting $v_i$ by increasing the domination level of its class on it ($v_i^{\omega_c}$, $c=\rho_j^c$) at the same time that it decreases the domination levels of other classes ($v_i^{\omega_c}$, $c \neq \rho_j^c$)). Since nodes corresponding to labeled pixels have constant domination levels, \eqref{eq:UpdateNodesPot} is not applied to them.

The particle strength changes according to the domination level of its class in the node it is visiting. Thus, at each iteration, a particle strength is updated as follows: $\rho_j^{\omega}(t) = v_i^{\omega_c}(t)$, where $v_i$ is the node being visited, and $c = \rho_j^c$.

When a node $v_i$ is being visited, the particle updates its class distance table as follows:
\begin{equation}\label{eq:UpdatePartDist}
    d_c^i(t+1) = \left\{
    \begin{array}{cl}
        d_c^q(t) + 1 & \mbox{if } d_c^q(t) + 1 < d_c^i(t) \\
        d_c^i(t) & \mbox{otherwise}
    \end{array}\right.,
\end{equation}
where $d_c^q(t)$ is the distance from the previous visited node to the closest labeled node of the particle class, and $d_c^i(t)$ is the current distance from the node being visited to the closest labeled node of the particle class. Notice that particles have no knowledge of the graph connection patterns. They are only aware of which are the neighbors of the node they are visiting. Unknown distances are discovered dynamically as the particles walk and distances are updated as particles naturally find shorter paths to the nodes.

At each iteration, each particle $\rho_j$ chooses a node $v_i$ among the neighbors of its current node to visit. The probability of choosing a node $v_i$ is given by: a) the particle class domination on it, $v_i^{\omega_c}$, and b) the inverse of its distance to the closest labeled node from the particle class, $d_c^i$, as follows:
\begin{equation}\label{eq:ProbMov}
    p(v_i|\rho_j) = \frac{W_{qi}}{2\sum_{\mu=1}^{n}{W_{q \mu}}} +  \frac{W_{qi} v_i^{\omega_c} (1+d_c^i)^{-2}}{2\sum_{\mu=1}^{n}{W_{q\mu} v_\mu^{\omega_c}} (1+d_c^\mu)^{-2}} ,
\end{equation}
where $q$ is the index of the node being visited by particle $\rho_j$, $c$ is the class label of particle $\rho_j$, $W_{qi} = 1$ if there is an edge between the current
node and the node $v_i$, and $W_{qi} = 0$ otherwise. A particle will stay on the visited node only if, after applying \eqref{eq:UpdateNodesPot}, its class domination level is the largest on that node; otherwise, the particle is expelled and it goes back to the node it was before, staying there until the next iteration.

The stop criterion is defined as follows. Periodically, the highest domination level on each node is taken and their mean is calculated ($\langle v_i^{\omega_{m}} \rangle$, $m=\arg\max_c v_i^{\omega_c}$). This value usually has a quick increase in the first iterations, then it stabilizes at a high level and it starts oscillating slightly. At this moment, for each node $v_i$, if $v_i^{\omega_c}>0.9$, then the class $c$ is assigned to the corresponding pixel ($y(x_i)=c$). The remaining nodes (if any) will be labeled at a second phase.

The second phase is a quick iterative process, where each unlabeled pixel $x_i$ adjusts its corresponding $\mathbf{v_i^\omega}$ as follows:
\begin{equation}
\mathbf{v_i^\omega}(t+1) = \frac{1}{a} \sum_{j \in \eta} \mathbf{v_j^\omega}(t) \dist(x_i,x_j),
\end{equation}
where $\eta$ is the subset of $a$ adjacents pixels of $x_i$. $a=8$, except in the borders where no wraparound is applied. $\dist(x_i,x_j)$ is the function that returns the Euclidean distance between features $x_i$ e $x_j$, weighted by $\lambda$. Therefore, each unlabeled pixel receives contributions of the neighboring pixels, which are proportional to their similarity. The second phase ends when $\langle v_i^{\omega_{m}} \rangle$ stabilizes. Now unlabeled pixels finally receive their labels, $y(x_i) = \arg\max_c v_i^{\omega_c}(t)$.

\section{Building Networks for PCC}
\label{sec:pesos}

As explained in Section \ref{sec:particlemodel}, pixel features may be scaled by a vector of weights $\lambda$ in order to emphasize/deemphasize each feature to the upcoming network generation step. Increased segmentation accuracy by PCC is expected with a proper choice of weights. Therefore, it is desirable to find methods to automatically define $\lambda$.

In \cite{Breve2015IJCNN}, $\lambda$ was optimized using a genetic algorithm \cite{Goldberg1989,Mitchell1998}, but the segmentation accuracy, measured comparing the algorithm output with ground truth images segmented by humans, was used as the fitness function. This approach was acceptable as proof of concept, but in real-world segmentation tasks ground truth images are not available. Thus, in \cite{Breve2015WVC}, four methods were proposed to automatically adjust $\lambda$ based on the data distribution for each feature and each class, with only the user labeled pixels considered. That approach led to some mixed results, three of the four methods were able to increase PCC segmentation accuracy, when compared to the results achieved without weighting the features, in at least one of the three tested images. But none of the four methods increased PCC segmentation accuracy for all the three tested images, which were extracted from the Microsoft GrabCut database \cite{Rother2004}.

In this paper, a different approach is proposed. Instead of evaluating individual features before the network construction, networks are built with some candidate values for $\lambda$. The resulting candidate networks are evaluated using a proposed network index. Therefore, finding a good $\lambda$ becomes an optimization problem, where the proposed network index is maximized.

This approach has the advantage of considering all the features together, already weighted by the candidate $\lambda$. Therefore, individual features, which are not good discriminators alone and would be deemphasized in previous approaches, may be combined to produce a proper network for PCC segmentation.

The proposed network index $\phi$, to be maximized, is calculated by analyzing the edges between labeled nodes in the candidate network. It is defined as follows:
\begin{equation}
\label{eq:indexnoexp}
    \phi = \frac{z_i}{z_t},
\end{equation}
where $z_i$ is the amount of edges between two labeled nodes representing the same class, and $z_t$ is the total amount of edges between any labeled nodes, no matter which class they belong. Thus, $\phi$ is higher as the proportion of edges between nodes of the same class increases. Candidate networks with fewer edges between nodes representing different classes are desirable, since this is a clue that different classes data are well-separated in that network, making the PCC job easier.

Notice that theoretically $0 \leq \phi \leq 1$, but $\phi \approx 1$ in most practical situations, so the difference in $\phi$ for networks built with different $\lambda$ may be very small. Therefore, an improved index $\alpha$ is defined as:
\begin{equation}
\label{eq:index}
    \alpha = \left( \frac{z_i}{z_t} \right)^\sigma ,
\end{equation}
where
\begin{equation}
    \sigma = \frac{\ln(0.5)}{\ln(\Phi)},
\end{equation}
with $\Phi$ as the result of \eqref{eq:indexnoexp} when it receives a network built without any feature weighting, i.e., the same as if $\lambda = \{1, 1, \ldots, 1\}$. Notice that $0 \leq \alpha \leq 1$ with the differences in $\alpha$ for different choices of $\lambda$ being much easier to notice then in $\sigma$. $\alpha<0.5$ means that the choice of $\lambda$ is probably bad and may lead to PCC accuracy worse than when it is applied to the features without any weighting. $\alpha>0.5$ means the choice of $\lambda$ is probably effective. The higher $\alpha$ is, more appropriate the network is expected to be. So, $\alpha$ is maximized to find a proper network to feed PCC.

\begin{figure*}
\centering
\begin{tabular}{cc}
\subfloat[]{\includegraphics[width=6cm]{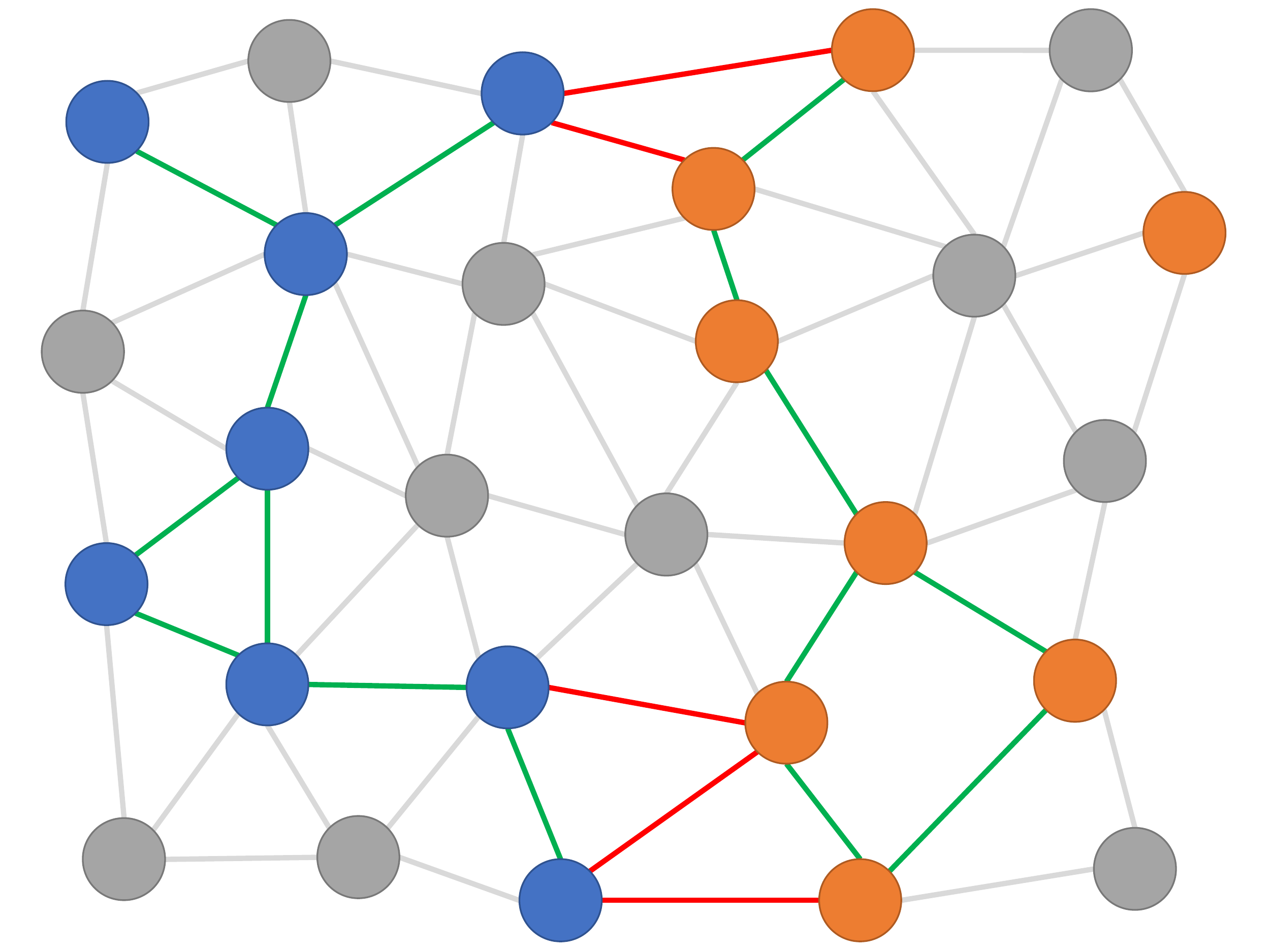} \label{fig:modelo}} &
\subfloat[]{\includegraphics[width=6cm]{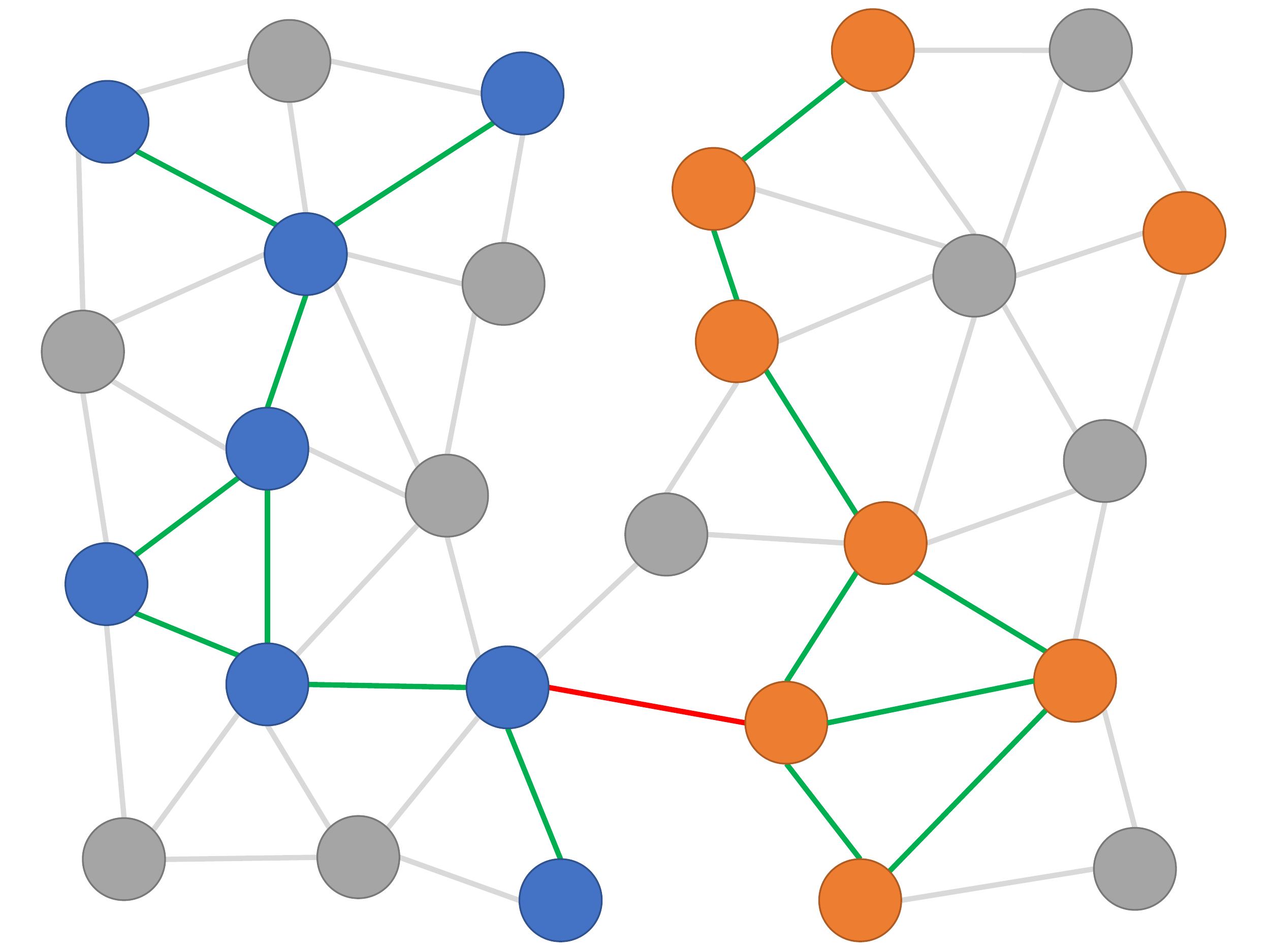} \label{fig:modelo2}}
\end{tabular}
\caption{Examples of candidate networks with $27$ nodes. Labeled nodes are colored in blue and orange. Unlabeled nodes are colored gray. (a) $15$ edges between nodes of the same class are represented in green, while $5$ edges between nodes of different classes are represented in red. (b) $16$ edges between nodes of the same class are represented in green, while a single edge between nodes of different classes is represented in red.}
\end{figure*}

Fig.~\ref{fig:modelo} shows a example of a candidate network. Suppose that it was built without any feature weighting. There are $27$ nodes, $8$ of them belong to the ``blue'' class, $8$ of them belong to the ``orange'' class, and the remaining $11$ nodes are unlabeled. There are $15$ edges (colored green) connecting nodes from the same class and $5$ edges (colored red) connecting nodes from different classes. Therefore, by applying \eqref{eq:indexnoexp}, $\phi = \frac{15}{20} = 0.75$. Then, $\sigma = 2.4094$. The index $\alpha$ for the same network will be $\alpha = \left( \frac{15}{20} \right)^{2.4094} = 0.5$.


Now, suppose that, during the optimization process, the network represented in Fig.~\ref{fig:modelo2} is built given a candidate $\lambda$. By applying \eqref{eq:index}, we have $\alpha = \left( \frac{16}{17} \right)^{2.4094} = 0.8641$. The higher $\alpha$ means that this network have higher class separability and it would probably allow PCC to achieve a higher classification accuracy then the network on Fig.~\ref{fig:modelo}


\section{Experiments}
\label{sec:experiments}

\begin{figure*}
\centering
\begin{tabular}{ccc}
\subfloat{\includegraphics[width=4.0cm]{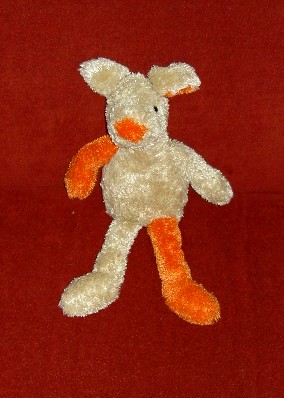}} &
\subfloat{\includegraphics[width=4.0cm]{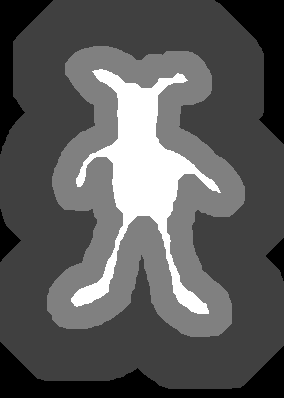}} &
\subfloat{\includegraphics[width=4.0cm]{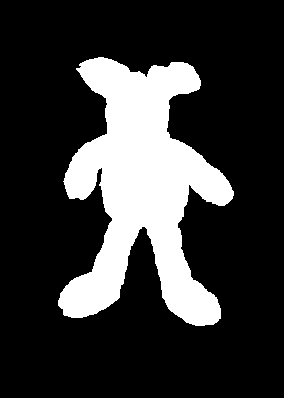}} \\
\subfloat{\includegraphics[width=4.0cm]{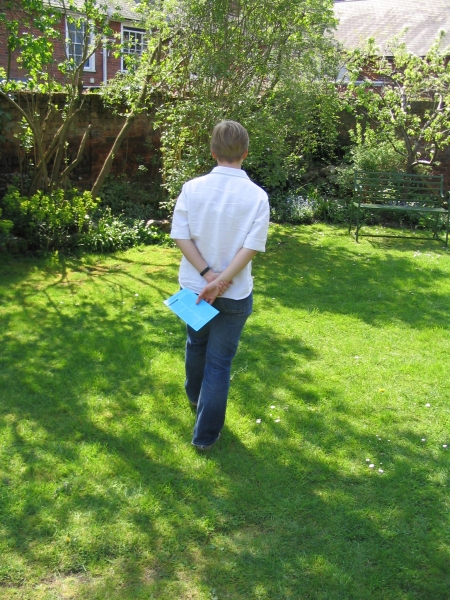}} &
\subfloat{\includegraphics[width=4.0cm]{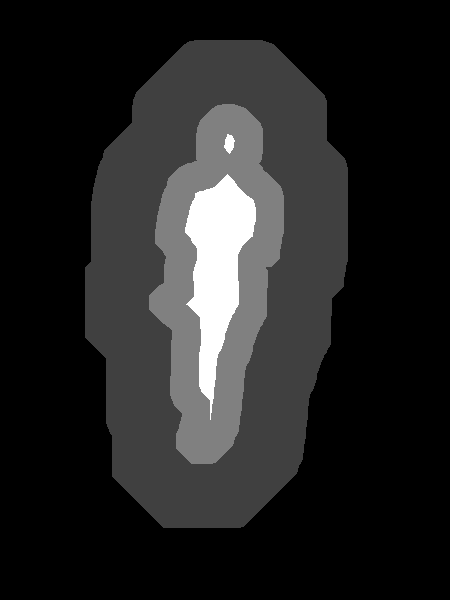}} &
\subfloat{\includegraphics[width=4.0cm]{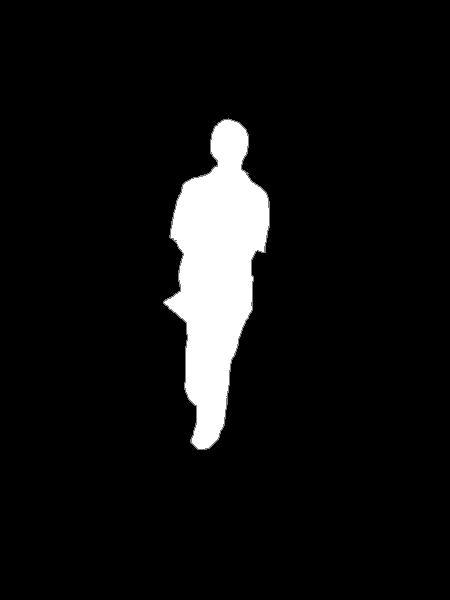}} \\
\subfloat{\includegraphics[width=4.0cm]{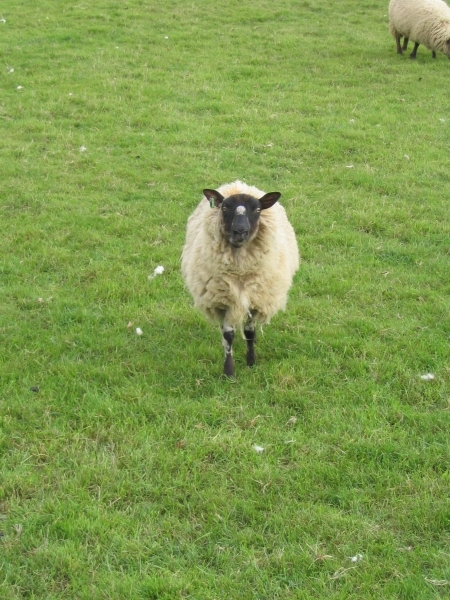}} &
\subfloat{\includegraphics[width=4.0cm]{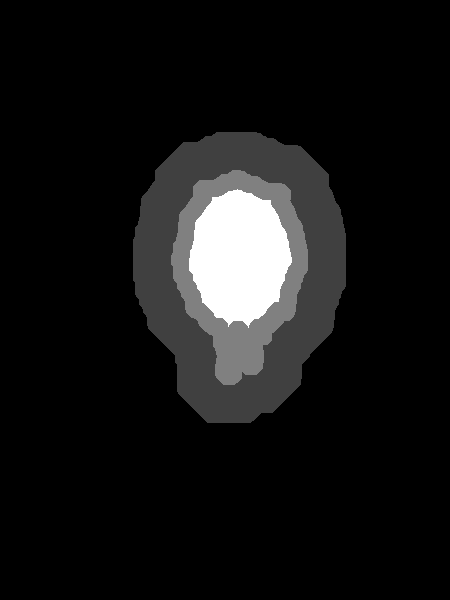}} &
\subfloat{\includegraphics[width=4.0cm]{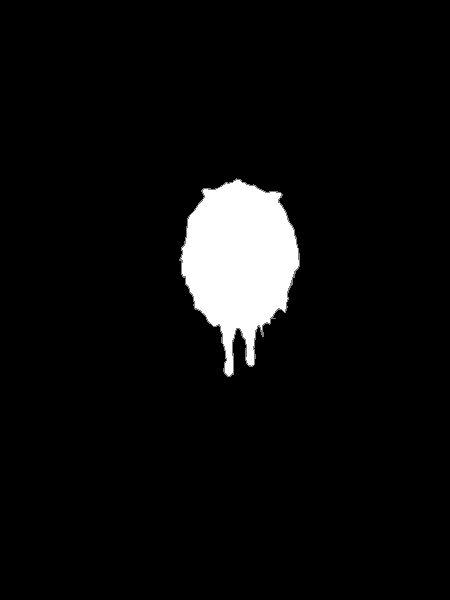}} \\
(a) & (b) & (c)
\end{tabular}
\caption{(a) Original images from the GrabCut dataset, (b) the \emph{trimaps} providing the seed regions, and (c) the original \emph{ground truth} images.}
\label{fig:grabcut-images}
\end{figure*}

In order to validate the proposed technique, three images were selected from the Microsoft GrabCut database \cite{Rother2004}. The selected images, their \emph{trimaps} providing seed regions, and the \emph{ground truth} images are shown on Fig.~\ref{fig:grabcut-images}. In the \emph{trimaps}, black (0) represents the background, which is ignored; dark gray (64) is the labeled background; light gray (128) is the unlabeled region, which labels will be estimated by the proposed method; and white (255) is the labeled foreground.

In the first experiment, networks were built for each image without any weighting and with different values for the parameter $k$. PCC was applied to each of them and the best segmentation accuracy result was taken for each image. These results are used as the baseline.

In the second experiment, for each image, the weight vector $\lambda$ was optimized using the genetic algorithm available in Global Optimization Toolbox of MATLAB, with its default parameters, while $k=100$ was kept fixed. Once the optimal $\lambda$ (based on the index $\sigma$) was found, networks with the optimal $\lambda$ and different values for the parameter $k$ were generated. PCC was applied to each of them and the best segmentation accuracy result was taken for each image as well.

\section{Results and Discussion}
\label{sec:results}
\begin{figure}
\centering
\begin{tabular}{cc}
\subfloat[Error: 1.89\%]{\includegraphics[width=4cm]{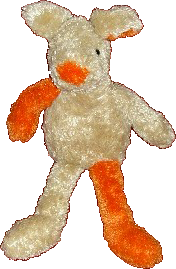}} &
\subfloat[Error: 1.86\%]{\includegraphics[width=4cm]{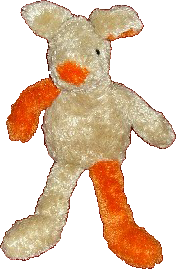}}
\end{tabular}
\caption{Teddy - Segmentation results achieved by PCC applied to: (a) networks built without feature weighting; (b) networks built with feature weights optimized by the proposed method}
\label{fig:res-teddy}
\end{figure}

\begin{figure}
\centering
\begin{tabular}{cc}
\subfloat[Error: 2.81\%]{\includegraphics[width=3cm]{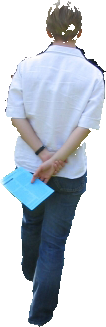}} &
\subfloat[Error: 1.67\%]{\includegraphics[width=3cm]{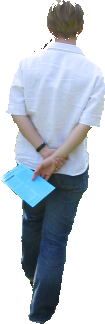}}
\end{tabular}
\caption{Person7 - Segmentation results achieved by PCC applied to: (a) networks built without feature weighting; (b) networks built with feature weights optimized by the proposed method}
\label{fig:res-person7}
\end{figure}

\begin{figure}
\centering
\begin{tabular}{cc}
\subfloat[Error: 2.90\%]{\includegraphics[width=3.5cm]{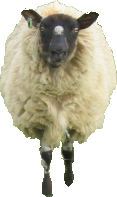}} &
\subfloat[Error: 2.04\%]{\includegraphics[width=3.5cm]{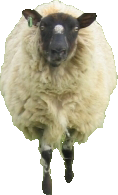}}
\end{tabular}
\caption{Sheep - Segmentation results achieved by PCC applied to: (a) networks built without feature weighting; (b) networks built with feature weights optimized by the proposed method}
\label{fig:res-sheep}
\end{figure}

The experiments described in Section \ref{sec:experiments} were applied on the three images shown on Fig.~\ref{fig:grabcut-images}. The best segmentation results achieved with the PCC applied to the networks without feature weighting and to the networks with the optimized weights are shown on Figs.~\ref{fig:res-teddy}, \ref{fig:res-person7}, and \ref{fig:res-sheep}. Error rates are computed as the fraction between the amount of incorrectly classified pixels and the total amount of unlabeled pixels (light gray on \emph{trimaps} images). Notice that \emph{ground truth} images have a thin contour of gray pixels, which corresponds to uncertainty, i.e., they received different labels by the different persons who did the manual classification. These pixels are not computed in the classification error.

Segmentation error rates are also summarized on Table~\ref{tab:error}. By analyzing the results we notice that the feature weight optimization using the proposed method lead to lower segmentation error rates on the three tested images, showing its effectiveness.

\begin{table}
  \centering
  \caption{Segmentation error rates when PCC is applied to networks built without feature weighting (baseline) and to networks built with feature weights optimized by the proposed method}
  \includegraphics[width=8cm]{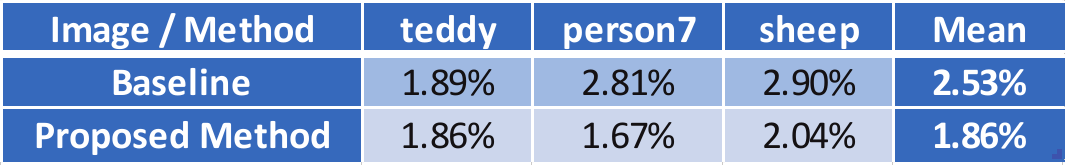}
  \label{tab:error}
\end{table}

The optimized indexes $\sigma$ found for each image were $1.0$ in all scenarios (32-bit float precision), which means they would probably improve the segmentation results, as they actually did. The networks generated for ``teddy'' easily reached $\sigma=1.0$, as more than half of the random selected weights would lead to $\sigma=1.0$. This explains why the first random generated weights (first individual) were returned by the genetic algorithm. On the other hand, ``person7'' and ``sheep'' took $40$ and $164$ generations, respectively, to finally reach $\sigma=1.0$. Each generation has $200$ individuals. The optimized features weights ($\lambda$) are shown on Table~\ref{tab:lambda}.

In the selected images, the row and the column of the pixels clearly are the most important features. Though the other features got lower weights in mean, the proper weights for each image were important to provide the decrease in classification error.


\begin{table}
  \centering
  \caption{Feature weights optimized by the proposed method}
  \includegraphics[width=8cm]{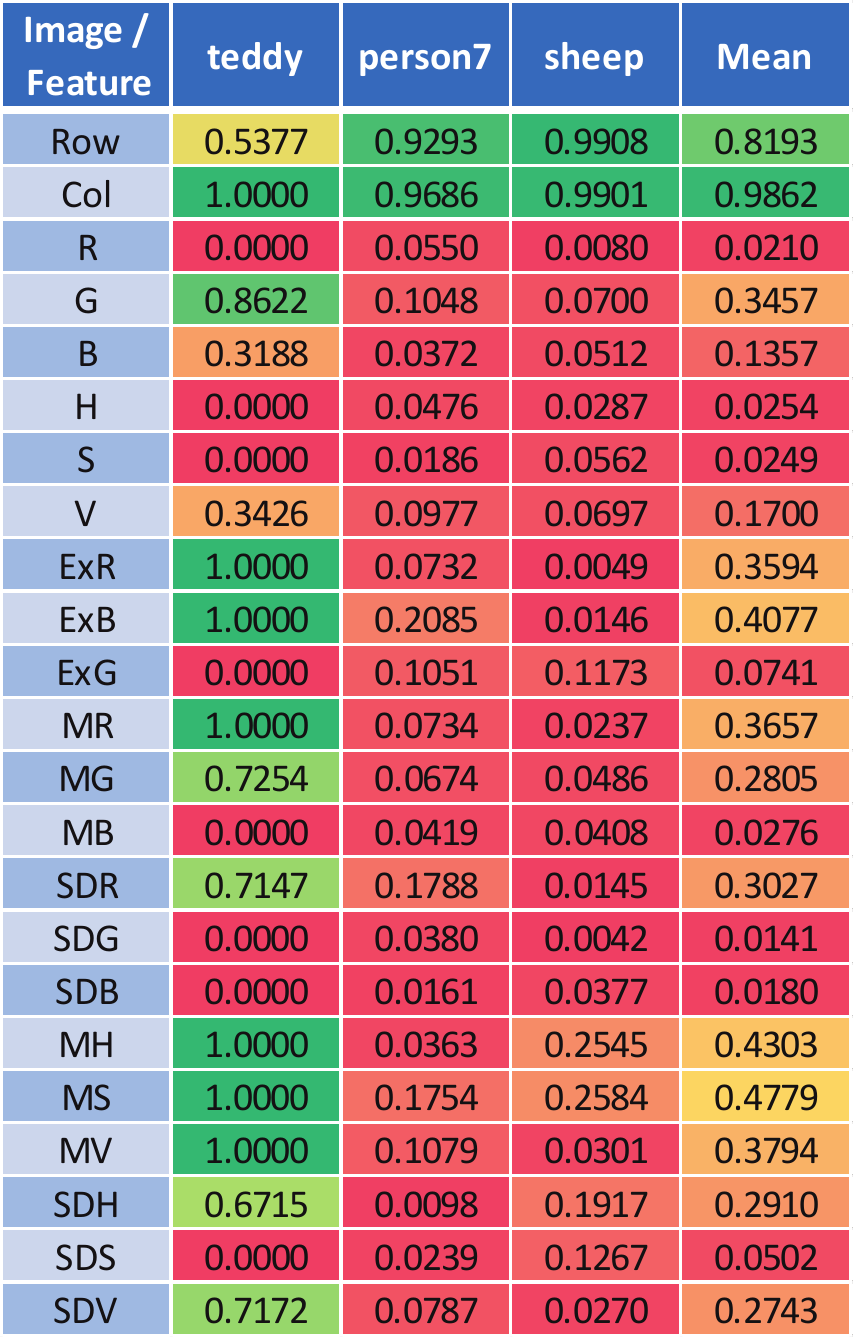}
  \label{tab:lambda}
\end{table}

\section{Conclusion}
\label{sec:conclusions}

In this paper, a new approach to build networks representing image pixels was proposed. The networks are used in the image segmentation task, using the semi-supervised learning method known as particle competition and cooperation (PCC). The approach consists in optimizing a proposed index which is calculated for each candidate network. The optimization process automatically calculates weights for the features which are extracted from the image to be segmented.

Computer simulations with some real-world images show that the proposed method is effective in improving segmentation accuracy, lowering pixel classification error. As future work, the method will be applied on more images and using more features, searching for some pattern on the images and the corresponding optimized weights. The index may also be improved to provide even better networks to feed PCC and further increase segmentation accuracy. The optimized feature weights might be used on similar images. Features with low weight might be excluded to improve execution time and segmentation accuracy. Finally, the method may be applied to images with less labeled pixels, like ``scribbles'' instead of ``trimaps'', since PCC is a semi-supervised method and does not require so many labeled data points.

\section*{Acknowledgment}

The author would like to thank the S\~{a}o Paulo Research Foundation - FAPESP (grant \#2016/05669-4) and the National Counsel of Technological and Scientific Development - CNPq (grant \#475717/2013-9) for the financial support.

%
%
\bibliographystyle{splncs03}
\bibliography{net4pcc}

\end{document}